\documentclass{article}

\usepackage[preprint]{neurips_2025}

\usepackage[utf8]{inputenc} 
\usepackage[T1]{fontenc}    
\usepackage[colorlinks=true, allcolors=blue]{hyperref}
\usepackage{url}            
\usepackage{booktabs}       
\usepackage{amsfonts}       
\usepackage{nicefrac}       
\usepackage{microtype}      
\usepackage{xcolor}         

\usepackage{graphicx}
\usepackage{amsmath}
\usepackage{amssymb}
\usepackage{subcaption}
\usepackage{algorithm}
\usepackage{algpseudocode}

\usepackage{tikz}
\usepackage{pgfplots}
\pgfplotsset{compat=1.18} 

\usepackage{filecontents}

\usepackage{tabularx}
\usepackage{ragged2e} 

\workshoptitle{Published in the proceedings of the 39th Conference on Neural Information Processing Systems (NeurIPS 2025) Workshop: Scaling Environments for Agents (SEA).}

\title{Co-Evolving Complexity: An Adversarial Framework for Automatic MARL Curricula}

\author{%
  Brennen A. Hill\\
  Department of Computer Science\\
  University of Wisconsin-Madison\\
  Madison, WI 53706 \\
  \texttt{bahill4@wisc.edu} \\
}

\begin{document}

\maketitle

\begin{abstract}
The advancement of general-purpose intelligent agents is intrinsically linked to the environments in which they are trained. While scaling models and datasets has yielded remarkable capabilities, scaling the complexity, diversity, and interactivity of environments remains a crucial bottleneck. Hand-crafted environments are finite and often contain implicit biases, limiting the potential for agents to develop truly generalizable and robust skills. In this work, we propose a paradigm for generating a boundless and adaptive curriculum of challenges by framing the environment generation process as an adversarial game. We introduce a system where a team of cooperative multi-agent defenders learns to survive against a procedurally generative attacker. The attacker agent learns to produce increasingly challenging configurations of enemy units, dynamically creating novel worlds tailored to exploit the defenders' current weaknesses. Concurrently, the defender team learns cooperative strategies to overcome these generated threats. This co-evolutionary dynamic creates a self-scaling environment where complexity arises organically from the adversarial interaction, providing an effectively infinite stream of novel and relevant training data. We demonstrate that with minimal training, this approach leads to the emergence of complex, intelligent behaviors, such as flanking and shielding by the attacker, and focus-fire and spreading by the defenders. Our findings suggest that adversarial co-evolution is a powerful mechanism for automatically scaling environmental complexity, driving agents towards greater robustness and strategic depth.
\end{abstract}

\section{Introduction}

The development of intelligent agents, especially those leveraging Large Language Models (LLMs), has underscored the foundational role of environments in cultivating sophisticated behaviors \citep{wang2019poet}. Environments are not merely passive arenas for evaluation; they are the interactive substrate from which agents learn adaptive behavior, complex reasoning, and long-term planning. The trajectory of progress in machine learning has been marked by scaling laws: increasing model size, dataset volume, and computational power has unlocked emergent capabilities \citep{Silver2017Mastering}. We posit that a similar principle applies to agent development, where scaling the structure, fidelity, and diversity of environments is a critical vector for advancing agent intelligence.

Recent breakthroughs in end-to-end reinforcement learning (RL) have made it feasible to train agents through sustained environmental interaction, moving beyond the limitations of imitation learning from static datasets \citep{Schulman2017Proximal}. This shift places a greater demand on the environments themselves. To foster general-purpose autonomy, we require environments that are not only richly interactive but also perpetually novel, preventing agents from overfitting to a fixed set of scenarios. The manual design of such environments is an intractable task, as it requires immense human effort and is subject to the designers' inherent biases and limited imagination.

This paper addresses the challenge of scaling environmental complexity through a novel framework: \textbf{Learned Adversarial Procedural Generation for Multi-Agent Curricula}. We reframe the problem of environment design as a two-player game between a generative \textit{Attacker} and a team of cooperative \textit{Defender} agents. The Attacker's goal is to learn a policy for procedurally generating sequences of hostile units (i.e., worlds or challenges) that can defeat the Defenders. Simultaneously, the multi-agent Defender team learns a cooperative policy to survive the Attacker's generated worlds for as long as possible.

This adversarial dynamic creates a natural and automatic curriculum. As the Defenders improve, the Attacker is incentivized to generate more sophisticated and complex challenges to remain competitive. This, in turn, forces the Defenders to develop more robust and coordinated strategies. To further drive this complexity, we designed the environment to have a combinatorially large action and state space. The defenders are not identical; each is assigned a unique role with different special abilities. The attacker, in turn, has a wide range of traits it can assign to the units it generates. In much of RL, problems are simple toy examples that do not extend beyond their initial experiment. By intentionally creating a more complex interaction with a vast space of possible environments and defender actions, we work to address that issue and create a more robust training paradigm. The result is a co-evolutionary arms race where the environment, embodied by the Attacker, continuously adapts to challenge the learning agents, effectively generating an infinite stream of increasingly difficult worlds.

Our primary contributions are:
\begin{enumerate}
    \item We present a system architecture for co-evolving a generative adversarial agent and a team of cooperative agents. The adversary's role is to procedurally generate environmental challenges, creating an open-ended learning process.
    \item We demonstrate that this adversarial framework leads to the rapid emergence of complex and intelligent strategies in both the generative Attacker and the cooperative Defender team.
    \item We provide qualitative and quantitative evidence of emergent behaviors, such as the Attacker learning to flank and shield its units, and the Defenders learning to coordinate their movements and focus fire, behaviors which were not explicitly programmed.
    \item We argue that this paradigm serves as a powerful method for scaling environments for agent training, shifting the focus from hand-crafting content to designing the rules of a self-perpetuating, complexity-generating system.
\end{enumerate}

The remainder of this paper is structured as follows: Section 2 reviews related work in multi-agent reinforcement learning, procedural content generation, and adversarial learning. Section 3 details the formal problem setting and our proposed system architecture. Section 4 describes the implementation, including the agent models and training regime. Section 5 presents our experimental results, focusing on the emergent behaviors. Section 6 discusses the implications of our findings and the limitations of the current work. Finally, Section 7 concludes with a summary of our contributions.

\section{Related Work}

Our research is situated at the intersection of three key areas: Multi-Agent Reinforcement Learning (MARL), Procedural Content Generation (PCG), and the use of adversarial dynamics to create automatic curricula.

\subsection{Multi-Agent Reinforcement Learning (MARL)}
MARL extends reinforcement learning to scenarios with multiple interacting agents. A central challenge in MARL is non-stationarity: from any single agent's perspective, the environment is constantly changing as other agents adapt their policies \citep{Busoniu2008Comprehensive}. This makes learning unstable. Our system embraces this non-stationarity, leveraging it as the primary driver of learning for both the Attacker and the Defenders.

MARL problems can be categorized as cooperative, competitive, or mixed-motive \citep{Zhang2019Selective}. Our work features a mixed structure: the Defender team is fully cooperative, while the relationship between the Defender team and the Attacker is fully competitive, forming a game that is close to zero-sum. The formal framework for such interactions is the Partially Observable Markov Game (POMG), where agents must make decisions based on incomplete information about the true game state \citep{Hansen2004Dynamic, liu2022sample}. In our setup, the Defenders have only partial observability of the Attacker's internal state, not seeing the Attacker's energy reserves and policy.

A dominant paradigm in modern MARL is Centralized Training with Decentralized Execution (CTDE) \citep{deWitt2020Is}. In CTDE, agents use global information (e.g., a shared value function) during training to stabilize learning but act based only on their local observations during execution. Proximal Policy Optimization (PPO) \citep{Schulman2017Proximal} has proven surprisingly effective in cooperative MARL settings when adapted to this paradigm (e.g., MAPPO), challenging the notion that on-policy methods are too sample-inefficient \citep{Yu2022Surprising}. This body of work provides strong justification for our choice of PPO as the learning algorithm for the cooperative Defender team and the competitive Attacker.

\subsection{Procedural Content Generation (PCG)}
Procedural Content Generation refers to the algorithmic creation of game content. Traditional PCG methods are often constructive or search-based. A more recent paradigm is PCG via Machine Learning (PCGML), where models are trained on existing content to generate new, similar content \citep{Summerville2018Procedural}. For example, models can learn to blend existing levels to create novel combinations \citep{Guzdial2018Learning}. However, PCGML is fundamentally imitative and its creative potential is bounded by its training data.

To overcome this limitation, PCG via Reinforcement Learning (PCGRL) was introduced, framing content generation as an RL problem where an agent learns to iteratively modify a level to maximize a reward function based on desired properties like playability \citep{Khalifa2020PCGRL}. This approach is inventive rather than imitative, as it can discover novel content through exploration. Our work builds directly upon this idea, but instead of using a static, hand-crafted reward function, the reward signal for our generative Attacker is derived dynamically from the performance of another learning agent (the Defender team).

\subsection{Adversarial Learning and Automatic Curricula}
The core mechanism of our system is the adversarial dynamic between the generator and the solvers. This concept has deep roots in machine learning, most notably in Generative Adversarial Networks (GANs). In the context of RL, adversarial self-play has been shown to be a powerful engine for generating complexity and achieving superhuman performance without human data, as exemplified by AlphaGo and AlphaZero \citep{Silver2016Mastering, Silver2017Mastering}. Similarly, competitive multi-agent environments have been shown to produce a natural curriculum, leading to the emergence of complex skills and strategies as agents continually adapt to one another \citep{Bansal2018Emergent, Tampuu2017Multiagent, Narvekar2020Curriculum}.

The explicit use of an adversary for PCG was explored by Volz et al. \citep{Volz2021Adversarial} and Gisslén et al. \citep{gisslen2021adversarial}, who proposed a Generator-Solver framework where the generator is rewarded for creating challenging but solvable levels for a single solver agent. Our work extends this adversarial PCG paradigm in several critical dimensions. We transition from a single-solver setting to a multi-agent cooperative team, elevating the task from solving static puzzles to developing dynamic, coordinated strategies against a learning adversary. Second, our generator operates at a more fundamental level with fine-grained control over the challenge. We shift the focus from generating solvable static environments to orchestrating a dynamic, self-scaling curriculum.

This process of co-evolution, where agents and their environments develop in tandem, has been identified as a powerful method for open-ended learning. The POET algorithm, for instance, co-evolves a population of environments and agent policies, leading to the continual generation of novel and complex challenges \citep{wang2019poet}. Other work has explored co-evolving an agent's morphology alongside its environment \citep{Ao2023Curriculum}. Our system can be seen as a specific instantiation of this broader principle, using a competitive game to drive the co-evolution of environmental challenges (from the Attacker) and solving policies (from the Defenders). This dynamic automatically generates goals of appropriate difficulty, a key principle in automatic curriculum generation \citep{Florensa2018Automatic}. Finally, our use of PPO is further supported by its extensibility for training policies robust to adversarial perturbations \citep{Wu2021Adversarial, Zhang2020Robust}.

\section{System Design and Methodology}

We formalize our system as a two-team, nearly zero-sum, partially observable Markov game (POMG) \citep{Hansen2004Dynamic}. The game consists of Team D, a set of $N=4$ cooperative Defender agents, and Team A, a single adversarial Attacker agent.

\subsection{Environment}
The game takes place on a discrete 2D grid, representing a board with 10 lanes (x-axis) and 30 tiles of depth (y-axis). The Defender agents are constrained to the first four rows ($y \in [0, 3]$), while the Attacker operates from the far end of the board ($y=29$). Time proceeds in discrete timesteps. The Defenders win by surviving, while the Attacker wins if a unit reaches the bottom edge or if any Defender is defeated.

\begin{figure}[h]
    \centering
    \includegraphics[width=0.4\linewidth]{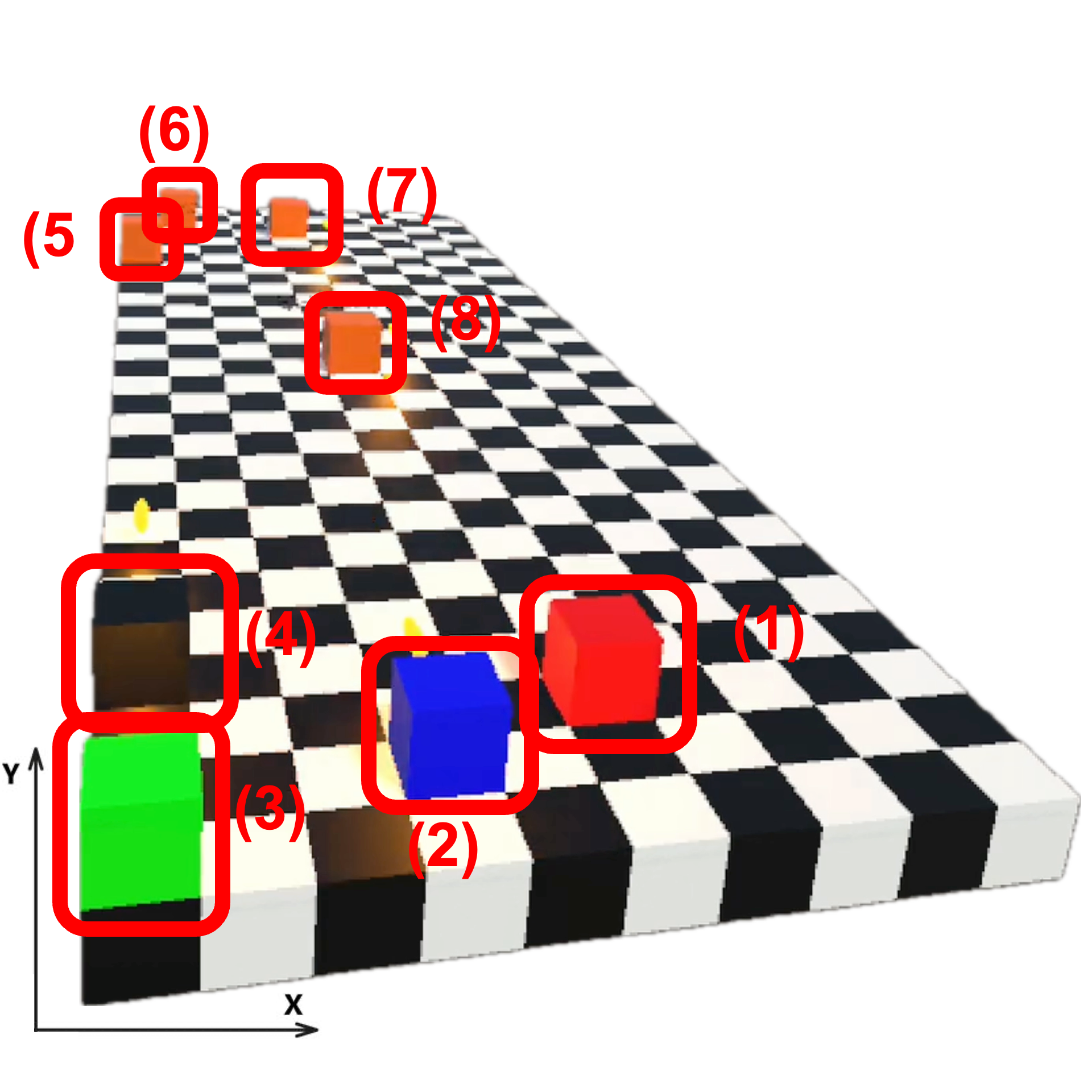}
    \caption{The game environment. The four Defender agents (1), (2), (3), and (4) can only move horizontally. The arbitrary number of orange Units (5), (6), (7), and (8) generated by the Attacker move vertically from the top of the board downwards.}
    \label{fig:environment}
\end{figure}

\subsection{Agents and Actions}

\subsubsection{The Defender Team}
The Defender team consists of four agents, each assigned one of four unique roles, detailed in Table \ref{tab:defender_roles}. Each Defender has a persistent state defined by its position $(x_i, z_i)$, current health (max 100), and current energy (max 1000), which replenishes at 1 unit/tick. At each tick, each Defender $i \in \{1, ..., 4\}$ chooses an action from the discrete action space detailed in Table \ref{tab:defender_actions}.

\begin{table}[h]

\caption{Defender agent action space and energy costs.}

\label{tab:defender_actions}

\centering

\begin{tabular}{@{}lll@{}}

\toprule

Action & Energy Cost     & Description                                       \\ \midrule

Move Left/Right & 5       & Change the agent's x-coordinate by -1 or +1.      \\

Shoot & 10              & Fire a projectile down the current lane.          \\

Heal & 50               & Restore a portion of its own health.              \\

Special Ability & 200     & A powerful, role-specific action.                 \\

Do Nothing & 0          & Conserve energy.                                  \\ \bottomrule

\end{tabular}

\end{table}

\subsubsection{The Attacker (Generative) Agent}
The Attacker agent's role is to procedurally generate challenges. Its state is defined by its current and maximum energy, which slowly increases over time. At each tick, the Attacker chooses an action $a^A$: generate a unit with specific parameters or do nothing to conserve energy. The parameter vector $\theta$ that defines a unit's attributes is the core of the procedural generation, creating a vast combinatorial space of possible enemies, detailed in Table \ref{tab:unit-parameters}.

\begin{table}[h]
\caption{Parameter space for the procedurally generated units by the Attacker agent. The agent can choose from a discrete set of values within the specified ranges to define a unit's attributes.}
\label{tab:unit-parameters}
\centering
\begin{tabular}{ll}
\textbf{Category} & \textbf{Parameter \& Range} \\
\midrule
Placement & Lane $x \in [0, 9]$ \\
Core Stats & Health (1-15), Damage (1-5), Speed (1-5), Range (1-25) \\
Special Attributes & Regeneration (0-3 health/tick), Leech (0-5 health on attack) \\
Defenses & Physical Defense (0-5), Magic Defense (0-5) \\
Offenses & Physical and (0-5), Magic Penetration (0-5), Type (Physical/Magic) \\
\bottomrule
\end{tabular}
\end{table}

The energy cost of generating a unit is a superlinear, multiplicative function of its parameters $\theta$. More powerful units are exponentially more expensive, forcing the Attacker to make strategic trade-offs between quantity and quality.

\subsubsection{Unit Behavior}
Generated units are not controlled by the Attacker. They follow a hard-coded behavior: move forward (decrease y-coordinate) each tick. If a Defender is in the same lane and within range, the unit stops moving and attacks. This ensures challenge complexity arises from the Attacker's generative choices, not from complex unit AI.

\subsection{Game Dynamics and Objectives}
An episode begins with four Defenders and the Attacker. The episode ends when one of two termination conditions is met: (1) An enemy unit reaches the Defenders' baseline ($y < 0$), or (2) Any Defender's health drops to zero. If either occurs, the Attacker wins. The Defenders' objective is to survive as long as possible; the Attacker's is to win as quickly as possible.

\section{Experiments and Results}

The primary goal was to investigate whether intelligent, complex, and adaptive strategies would emerge from the adversarial co-evolutionary process. Success was measured by the qualitative and quantitative richness of the observed behaviors after 500 episodes, compared against a baseline where both sides selected actions uniformly at random.

\subsection{Baseline Comparison}
The baseline agents exhibited no intelligent behavior. The random attacker generated units with arbitrary parameters at random locations and times. The random defenders moved without purpose, failing to engage units or wasting energy. Episodes were brief, with defenders being quickly overwhelmed.

\subsection{Emergent Behaviors}
Intelligent strategies had clearly emerged after 500 episodes. The co-evolutionary pressure forced both sides to develop sophisticated tactics to counter each other.

\subsubsection{Emergent Attacker Strategies}
The Attacker learned to move beyond simply generating strong enemies and began to exhibit temporal and spatial reasoning.

The Attacker learned to move beyond simply generating strong enemies and began to exhibit temporal and spatial reasoning. It developed several distinct tactics, including: the \textbf{Rusher strategy}, where it generated units with very high speed and minimal other stats to race past defenders; the \textbf{Tandem strategy} (Figure \ref{fig:tandem}), a notable tactic of generating a high-health unit to act as a shield for a high-damage unit directly behind it in the same lane; and the \textbf{Flanking strategy} (Figure \ref{fig:flanking}), where it exploited the Defenders' limited mobility by generating simultaneous threats on opposite sides of the board.

\subsubsection{Emergent Defender Strategies}
In response, the Defender team developed coordinated behaviors. Despite a shared policy and no explicit communication channel, their actions became implicitly coordinated.

In response, the Defender team developed coordinated behaviors. Despite a shared policy and no explicit communication channel, their actions became implicitly coordinated. For example, they learned \textbf{Cooperative Spreading (Figure \ref{fig:spread})}, where, when faced with multiple threats (such as from Flanking), defenders learned to spread out to cover the relevant lanes. Conversely, they also learned \textbf{Cooperative Focusing (Figure \ref{fig:focus})}, where multiple defenders would converge on the same lane to concentrate firepower on a single, high-priority threat (such as from the Tandem strategy).

\subsection{Quantitative Analysis}
To ground our observations, we quantified the frequency of emergent strategies by defining and watching for four signature behaviors detailed in Table \ref{tab:strategy_definitions}.  quantitative metrics reported in this section are the average values obtained from 100 independent runs to ensure statistical reliability.

Table \ref{tab:strategy_stats} presents the statistics for these strategies after 500 training episodes, averaged across 100 independent runs, comparing the trained agents to the random baseline. The difference is stark. The trained agents' average episode length was over four times longer than the baseline (83 steps vs. 19), demonstrating a vastly superior ability to survive. This increased survival time is directly attributable to the adoption of coherent strategies.

As shown in Table \ref{tab:strategy_stats}, the trained Attacker employed the Tandem and Flanking strategies in over 98\% and 94\% of episodes, respectively. These were not rare occurrences but the core of its learned policy. Similarly, the Defender team utilized Cooperative Spreading and Focusing in 92.6\% and 81.4\% of episodes. In contrast, the random baseline agents triggered these strategic patterns at rates below 11\%, consistent with chance occurrences in a short episode. The average uses per episode show that the trained agents repeatedly and deliberately execute these tactics, whereas the random agents barely perform them once across ten episodes. This data provides strong quantitative validation that the adversarial process did not just improve agent performance but induced the learning of specific, recognizable, and effective multi-agent tactics.

\begin{table}[h]
\centering
\caption{Frequency of Emergent Strategies After 500 Episodes, averaged across 100 runs.}
\label{tab:strategy_stats}
\begin{tabular}{llcc}
\toprule
\textbf{Agent Type} & \textbf{Strategy Metric} & \textbf{Trained Agents} & \textbf{Random Baseline} \\
\midrule
\multicolumn{4}{c}{\textit{\textbf{Defender Cooperative Strategies}}} \\
\midrule
Defender & Cooperative Spreading (Avg. Uses/Ep) & 3.61 & 0.0104 \\
         & Cooperative Spreading (Usage Rate)   & 92.6\% & 0.837\% \\
\cmidrule{2-4}
         & Cooperative Focusing (Avg. Uses/Ep)  & 2.97 & 0.00906 \\
         & Cooperative Focusing (Usage Rate)    & 81.4\% & 0.548\% \\
\midrule
\multicolumn{4}{c}{\textit{\textbf{Attacker Generative Strategies}}} \\
\midrule
Attacker & Flanking (Avg. Uses/Ep)              & 4.85 & 0.128 \\
         & Flanking (Usage Rate)                & 94.0\% & 10.3\% \\
\cmidrule{2-4}
         & Tandem (Avg. Uses/Ep)                & 8.01 & 0.0919 \\
         & Tandem (Usage Rate)                  & 98.2\% & 6.73\% \\
\midrule
\multicolumn{2}{l}{\textbf{Avg. Episode Length (steps)}} & \textbf{83} & \textbf{19} \\
\bottomrule
\end{tabular}
\end{table}

\subsection{Ablation Study: The Necessity of Co-Evolution}

To isolate the impact of the adversarial dynamic, we conducted two additional experiments where one agent was trained against a non-learning, random opponent for 500 episodes. The results, summarized in Table \ref{tab:ablation_stats}, underscore that co-evolution is the primary driver of strategic complexity.

First, we trained the Defender team against a perpetually random Attacker. The Defenders survived significantly longer against a random generator, with an average episode length of 216 steps. They rarely employed cooperative strategies, using Spreading at a rate of $13.2\%$ and Focusing at $9.30\%$. Accounting for the longer episodes, the Defenders hardly improved from the random baseline. Qualitatively, many of these rare instances appeared more as random chance than intentional maneuvers. The challenge remained simple, and the learning plateaued. 

Conversely, we trained the Attacker against a team of random Defenders. The results were even more telling. The average episode length plummeted to just 14 steps, as the random Defenders offered virtually no resistance. Critically, the incentive to develop intelligent strategies vanished. The Flanking and Tandem strategies appeared at rates of only $13.7\%$ and $21.2\%$, respectively. Without a competent opponent to challenge it, the Attacker's policy failed to develop strategic depth, succeeding simple unit spawns. Together, these ablations provide strong evidence that it is the mutual, reciprocal adaptation, the co-evolutionary arms race, that generates the rich, emergent behaviors observed in our main experiment.

\begin{table}[h]

\centering

\caption{Ablation Study: Strategy Frequency When Training Against a Random Opponent, averaged over 100 independent runs.}

\label{tab:ablation_stats}

\begin{tabular}{llc}

\toprule

\textbf{Agent Type} & \textbf{Metric} & \textbf{Value} \\

\midrule

\multicolumn{3}{c}{\textit{\textbf{Trained Defender vs. Random Attacker}}} \\

\midrule

Defender & Avg. Episode Length (steps) & 216 \\

\cmidrule{2-3}

& Cooperative Spreading (Avg. Uses/Ep) & 0.188 \\

& Cooperative Spreading (Usage Rate) & 13.2\% \\

\cmidrule{2-3}

& Cooperative Focusing (Avg. Uses/Ep) & 0.113 \\

& Cooperative Focusing (Usage Rate) & 9.30\% \\

\midrule

\multicolumn{3}{c}{\textit{\textbf{Trained Attacker vs. Random Defender}}} \\

\midrule

Attacker & Avg. Episode Length (steps) & 14 \\

\cmidrule{2-3}

& Flanking (Avg. Uses/Ep) & 0.184 \\

& Flanking (Usage Rate) & 13.7\% \\

\cmidrule{2-3}

& Tandem (Avg. Uses/Ep) & 0.328 \\

& Tandem (Usage Rate) & 21.2\% \\

\bottomrule

\end{tabular}

\end{table}

\begin{figure}[h!]
    \centering
    \begin{tikzpicture}
        \begin{axis}[
            xlabel={Training Episode},
            ylabel={Average Episode Length (Ticks)},
            xmin=0, xmax=520,
            ymin=0, ymax=100,
            xtick={0, 100, 200, 300, 400, 500},
            ytick={0, 20, 40, 60, 80, 100},
            legend pos=north west,
            grid=major,
            grid style={dashed,gray!30},
            width=0.64\textwidth,
            height=0.4\textwidth,
            title={Defender Survival Time Over Training}, 
            ]
            \addplot[
                color=blue,
                smooth,
                thick
                ]
                coordinates {
                    (0.00, 18.00) (5.00, 23.11) (10.00, 25.05) (15.00, 20.05) (20.00, 30.25) (25.00, 31.78) (30.00, 32.39) (35.00, 40.13) (40.00, 39.19) (45.00, 41.16) (50.00, 36.72) (55.00, 45.32) (60.00, 45.18)
                    (65.00, 20.00) (70.00, 22.70) (75.00, 35.01) (80.00, 44.43) (85.00, 47.00) (90.00, 47.03) (95.00, 46.66) (100.00, 49.96) (105.00, 46.90) (110.00, 48.82) (115.00, 50.10)
                    (120.00, 26.38)
                    (125.00, 37.23) (130.00, 43.72) (135.00, 53.89) (140.00, 55.13) (145.00, 51.38) (150.00, 56.46) (155.00, 53.55)
                    (160.00, 38.20)
                    (165.00, 44.89) (170.00, 55.21) (175.00, 60.92) (180.00, 57.44) (185.00, 56.32) (190.00, 61.54) (195.00, 60.11) (200.00, 58.09) (205.00, 61.58) (210.00, 58.40) (215.00, 65.90) (220.00, 61.19)
                    (225.00, 36.05) (230.00, 43.15) (235.00, 56.95) (240.00, 65.41) (245.00, 62.98) (250.00, 69.72) (255.00, 68.62) (260.00, 70.39) (265.00, 70.48) (270.00, 68.55) (275.00, 71.58) (280.00, 65.35) (285.00, 66.64) (290.00, 65.86) (295.00, 68.53) (300.00, 69.46)
                    (305.00, 47.67)
                    (310.00, 58.18) (315.00, 64.19) (320.00, 70.25) (325.00, 72.75) (330.00, 69.93) (335.00, 75.62) (340.00, 70.20) (345.00, 77.89) (350.00, 76.56) (355.00, 72.36) (360.00, 71.20) (365.00, 78.06) (370.00, 77.57) (375.00, 78.12) (380.00, 78.84) (385.00, 73.63) (390.00, 76.27) (395.00, 74.70) (400.00, 81.04) (405.00, 79.49) (410.00, 74.51) (415.00, 65.73) (420.00, 75.06) (425.00, 78.53) (430.00, 82.12) (435.00, 81.73) (440.00, 84.07) (445.00, 81.10) (450.00, 78.62) (455.00, 83.71) (460.00, 84.43) (465.00, 83.17) (470.00, 85.19) (475.00, 83.30) (480.00, 83.87) (485.00, 83.44) (490.00, 80.55) (495.00, 81.54) (500.00, 81.25)

                };
            \legend{Defender Survival}
        \end{axis}
    \end{tikzpicture}
    \caption{An illustration of the average episode length (Defender survival time in ticks) over training Episode. The plot represents the general upward trend indicating skill improvement, alongside the oscillations suggesting an ongoing arms race where the Attacker discovers new strategies. This curve is representative of the observed dynamic rather than a plot of raw data from a single training run.}
    \label{fig:survival_time}
\end{figure}
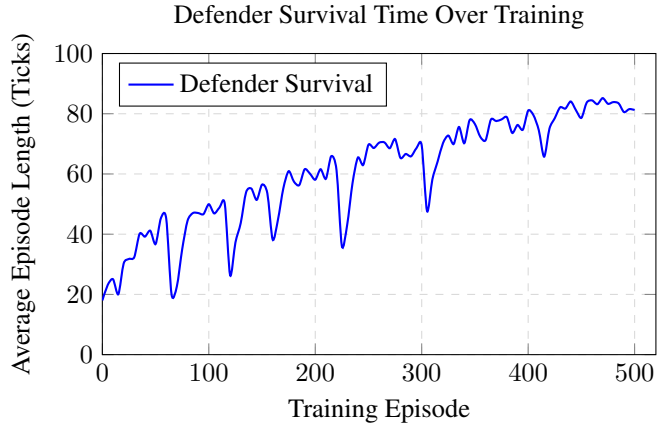

This strategic evolution is also reflected in the overall training dynamics. As shown in Figure \ref{fig:survival_time}, the average survival time of the defenders generally increased, indicating skill improvement. However, the curve is not monotonic; it exhibits significant oscillations. These dips often correspond to moments where the Attacker discovers a new, effective strategy that temporarily overcomes the Defenders' current policy. The Defenders then adapt, and the survival time climbs again. This oscillating pattern is characteristic of a co-evolutionary arms race and provides evidence of the ongoing adaptive process.

\section{Discussion}

Our results demonstrate that a co-evolutionary, adversarial framework can automatically generate a rich and adaptive curriculum for multi-agent systems. This section discusses the broader implications of this finding, situating it within the context of automatic curriculum generation, the nature of the emergent arms race, its impact on agent generalization, and its limitations.

\subsection{Adversarial Generation as a Self-Scaling Curriculum}
The findings strongly support the hypothesis that adversarial co-evolution is a potent mechanism for automatic curriculum generation. The Attacker agent, motivated to defeat the Defenders, is intrinsically driven to generate challenges at the frontier of the Defenders' capabilities. Challenges that are too easy do not yield rewards, while those that are truly impossible are prohibitively expensive in terms of energy, disincentivizing their creation. Consequently, the Attacker is naturally guided to probe and exploit specific weaknesses in the Defenders' current collective policy.

This process transforms the environment generator into a learned, adaptive loss function for the solver agents. Instead of training against a static dataset of challenges, the Defenders learn against a dynamic adversary whose sole purpose is to maximize their failure. This dynamic efficiently searches the combinatorially vast space of possible worlds (diverse units in Table \ref{tab:unit-parameters}) to find the small subset of configurations that are maximally informative for driving learning. The efficacy of this curriculum is quantitatively evident; trained agents adopt specific, complex tactics such as the Attacker's Tandem strategy (employed in 98.2\% of episodes) and the Defenders' Cooperative Spreading (92.6\%) (Table \ref{tab:strategy_stats}). This stands in stark contrast to our ablation studies, where training against a static, random opponent resulted in significantly less strategic depth (Table \ref{tab:ablation_stats}). The infinite stream of novel never-ending worlds is not just a theoretical construct; it is the emergent outcome of this adversarial dynamic, ensuring that the agents never exhaust their supply of relevant training data.

\subsection{The Nature of the Co-evolutionary Arms Race}
The oscillating performance curve seen in Figure \ref{fig:survival_time} is a hallmark of a co-evolutionary arms race, a dynamic akin to the Red Queen hypothesis in evolutionary biology, where species must constantly adapt simply to maintain their viability against evolving competitors \citep{van1973new}. The dips in Defender survival time likely correspond to moments where the Attacker discovers a new, effective strategy (e.g., a novel combination of unit parameters that bypasses the Defenders' current meta). This creates a strong learning signal for the Defender team, which must then adapt and develop a counter-strategy, leading to a subsequent rise in survival time.

This perpetual non-stationarity, driven by the arms race, is a desirable feature for open-ended learning. Our ablation studies confirm its necessity; when one side ceased to adapt, the strategic evolution of the other quickly stalled (Table \ref{tab:ablation_stats}). It prevents the agents from converging to a single, brittle policy. Instead of reaching a stable Nash equilibrium, the system navigates a continuous cycle of strategies and counter-strategies. For instance, the Attacker may favor Flanking, forcing the Defenders to master Spreading. A proficient Spreading defense then incentivizes the Attacker to pivot to a Tandem strategy, which in turn requires the Defenders to learn Focusing. This cycle prevents catastrophic forgetting of old skills, as any abandoned strategy may be re-exploited by the adversary, compelling agents to maintain a broad and robust repertoire of behaviors.

\subsection{Implications for Generalization and Robustness}
A primary goal in scaling environments is to produce agents that generalize to unseen situations. Our adversarial framework directly promotes generalization by design. Because the environment is actively hostile and non-stationary, the Defender agents cannot succeed by merely memorizing solutions to a fixed set of problems. They are forced to learn a more general, adaptive policy.

Furthermore, the Attacker functions as an automated red-teaming agent. It is trained to find the edge cases and blind spots in the Defenders' collective policy. This provides a far more efficient and exhaustive method for improving agent robustness than manual testing or curation of test cases. The system inherently generates a stress-testing suite that is always tailored to the agent's present capabilities, hardening the agent against a wide range of potential exploits. This suggests that such adversarial generation frameworks could become a standard component in pipelines for developing safe and reliable autonomous agents.

\subsection{Limitations and Future Work}
While promising, this work has several limitations that open avenues for future research. The training was conducted on consumer hardware for only 500 episodes; a longer training period would likely reveal even more sophisticated, multi-layered strategies.

A key direction for future work is the integration of Large Language Models (LLMs). An LLM could act as the generative Attacker, tasked with formulating high-level strategic goals (e.g., "create a pincer movement using fast units") which are then translated into specific unit generation actions. This would test the LLM's capacity for strategic reasoning in an interactive setting. Conversely, LLMs could be used by the Defender team for high-level planning or explicit communication, enabling more complex coordination.

Another promising avenue lies in scaling the complexity of the generator itself. The Attacker could be empowered to modify the environment's topology, place obstacles, or even design new types of units with unique, hard-coded behaviors. This concept also relates to tool-use; the Attacker's current action space is compositional, as it combines attributes (tools) to create a unit. Expanding this to a richer set of environmental tools would be a natural next step.

Finally, while we identified emergent strategies qualitatively and quantitatively, a deeper analysis of the learned policies is warranted. Techniques from explainable AI could be used to dissect the agents' decision-making processes, providing clearer insight into the mechanics of the co-evolutionary learning process. Exploring population-based training, where multiple species of Attackers and Defender teams co-evolve, could also lead to a richer and more diverse ecosystem of emergent strategies.

\section{Conclusion}

We have presented a system for multi-agent learning driven by adversarial procedural generation. By framing the interaction between a generative Attacker and a cooperative Defender team as a nearly zero-sum game, we successfully created a co-evolutionary dynamic that automatically scales environmental complexity in a targeted, adaptive manner. Our implementation demonstrates that this approach fosters the rapid emergence of intelligent, coordinated, and non-trivial strategies in both the generator and the solver agents, a finding supported by both qualitative observation and quantitative analysis.

This work contributes to the growing body of evidence that adversarial self-play is a powerful paradigm for generating complexity without human data. It offers a practical path forward for scaling environments to meet the demands of increasingly general and autonomous agents. By shifting the research focus from the manual design of static content to the architectural design of self-scaling, complexity-generating systems, we can create training methodologies that continuously challenge our agents, pushing them towards greater robustness, strategic depth, and general intelligence.

\medskip

{
\small
\bibliographystyle{plainnat}
\bibliography{main} 
}

\appendix

\appendix

\section{Environment Design Details}
\label{sec:appendix_environment}

This section provides additional details on the game's formal structure, reward functions, and the specific roles of the Defender agents.

\subsection{Formalism as a Partially Observable Markov Game}
The system is a POMG defined by the tuple $\langle I, S, \{A_i\}_{i \in I}, T, R, \{\Omega_i\}_{i \in I}, O \rangle$. Here, $I = \{A, D_1, D_2, D_3, D_4\}$ is the set of agents; $S$ is the global state space containing the positions, health, and energy of all agents and units; $A_i$ is the action space for agent $i$; $T(s'|s, \vec{a})$ is the state transition function; $R_i(s, \vec{a})$ is the reward function, which is nearly zero-sum such that $R_A \approx - \sum_{j=1}^{4} R_{D_j}$; $\Omega_i$ is the observation space for agent $i$; and $O(o_i|s, a_i)$ is the observation function. Each Defender observes its own state, the state of other defenders, and nearby units, but not the Attacker's energy. The Attacker observes the full game state.

\subsection{Reward Structure}
The reward functions are designed to incentivize the core objectives of each side.
\begin{itemize}
    \item \textbf{For the Defenders}, the reward signal includes a large negative reward for losing ($R_{loss} = -1.0$), a small positive reward for each tick survived ($R_{tick} = +0.001$), and a shaping reward for destroying an enemy unit ($R_{kill} = +0.05$).
    \item \textbf{For the Attacker}, the signal is symmetrical: a large positive reward for winning ($R_{win} = +1.0$), a small penalty per tick ($R_{tick} = -0.001$), and a shaping penalty for attempting to spawn a unit with insufficient energy ($R_{fail} = -0.03$).
\end{itemize}
This structure creates a strong competitive pressure, driving both sides to improve.

\subsection{Defender Role Specifications}
Each of the four Defender agents is assigned a unique role with distinct statistics and a powerful special ability, encouraging strategic differentiation within the cooperative team. The details are specified in Table \ref{tab:defender_roles}.

\begin{table}[h!]
  \caption{Detailed Defender role specifications, including damage type, base statistics, and unique special abilities.}
  \label{tab:defender_roles}
  \centering
  \begin{tabularx}{\linewidth}{l l l >{\RaggedRight}X}
    \toprule
    \textbf{Role} & \textbf{Damage Type} & \textbf{Base Statistics} & \textbf{Special Ability (Cost: 200 Energy)} \\
    \midrule
    \textbf{Mage} (Blue) &
    Magic &
    Low Phys. Def. \newline High Magic Def. &
    \textbf{\textit{Debuff Enemies}}: Removes all physical and magic defense from all active enemy units. \\
    \addlinespace
    \textbf{Healer} (Green) &
    Magic &
    Low Overall Stats &
    \textbf{\textit{Total Party Heal}}: Performs a large heal on all four friendly defenders. \\
    \addlinespace
    \textbf{Tank} (Black) &
    Physical &
    High Phys. Def. \newline Low Magic Def. &
    \textbf{\textit{Cannon}}: Deals massive area-of-effect physical damage to the densest cluster of enemies. \\
    \addlinespace
    \textbf{Sharpshooter} (Red) &
    Physical &
    High Phys. Pen. \newline Low Defenses &
    \textbf{\textit{Clear Lane}}: Deals very high damage to all enemy units in the Sharpshooter's current lane. \\
    \bottomrule
  \end{tabularx}
\end{table}

\section{Implementation Details}
\label{sec:appendix_implementation}

This section describes the technical implementation, including the agent architectures and the hyperparameters used for training.

\subsection{Training Environment and Agent Architectures}
We implemented our system using the Unity game engine and the ML-Agents Toolkit. Both agent types use a Multi-Layer Perceptron (MLP) architecture with two hidden layers of 128 neurons each using the ReLU activation function. The choice of a simple architecture was deliberate to emphasize that emergent complexity arises from environmental interaction rather than from an overly complex model.

\subsubsection{Defender Model}
The Defender model uses a 126-dimensional input observation vector, which includes the agent's own status (energy, health, position, role), the status of the other three defenders, and the attributes of up to 16 nearby enemy units. The output is a 6-node layer for the discrete action space, producing a probability distribution via a softmax function.

\subsubsection{Attacker (Generator) Model}
The Attacker model takes a 254-dimensional input observation vector, containing its own status (energy and max energy) and the full status of all defenders and up to 16 active units. Its complex action is modeled with a multi-branched output consisting of 13 separate heads, one for each unit parameter detailed in Table \ref{tab:unit-parameters}. A softmax is applied to each head independently, allowing the agent to learn a joint distribution over all unit parameters.

\subsection{Training Algorithm and Hyperparameters}
Both the Defender team and the Attacker are trained simultaneously using Proximal Policy Optimization (PPO) \citep{Schulman2017Proximal}. PPO is an on-policy, actor-critic algorithm known for its stability, making it a strong choice for this complex multi-agent setting \citep{Yu2022Surprising}. The four Defender agents are trained using a shared policy to encourage the development of cooperative strategies. The policies of the Attacker and Defenders are updated concurrently. 100 training runs were conducted for 500 episodes on a consumer-grade laptop (Intel Core i5-1035G7 CPU). An episode corresponds to a single game ending in a Defender loss. Key hyperparameters are listed in Table \ref{tab:hyperparams}.

\begin{table}[h!]
\centering
\caption{Training Hyperparameters for PPO.}
\label{tab:hyperparams}
\begin{tabular}{lc}
\toprule
\textbf{Hyperparameter} & \textbf{Value} \\
\midrule
Learning Rate ($\alpha$) & $3.0 \times 10^{-4}$ \\
Batch Size & 128 \\
PPO Epsilon ($\epsilon$) & 0.2 \\
Entropy Bonus ($\beta$) & $5.0 \times 10^{-4}$ \\
\bottomrule
\end{tabular}
\end{table}

\section{Emergent Strategy Details}
\label{sec:appendix_results}

This section provides the specific definitions used to quantify the emergent strategies discussed in the main paper, along with visualizations of these strategies in action.

\subsection{Strategy Definitions}
To ground our qualitative observations, we quantified the frequency of emergent strategies by defining four signature behaviors. These definitions, provided in Table \ref{tab:strategy_definitions}, were used to generate the statistics in Table \ref{tab:strategy_stats}.

\begin{table}[h!]
    \centering
    \caption{Definitions used for quantifying emergent strategies.}
    \label{tab:strategy_definitions}
    \begin{tabularx}{\linewidth}{l >{\RaggedRight}X}
        \toprule
        \textbf{Strategy} & \textbf{Quantifiable Definition} \\
        \midrule
        Cooperative Spreading & Triggered when no two defenders occupy the same lane for at least 5 consecutive timesteps. \\
        \addlinespace
        Cooperative Focusing & Triggered when at least three defenders occupy the same lane for at least 2 consecutive timesteps. \\
        \addlinespace
        Flanking & Triggered when the Attacker spawns units on both the far-left lanes (0 or 1) and far-right lanes (8 or 9) within a 2-timestep window. \\
        \addlinespace
        Tandem & Triggered when the Attacker spawns a unit into a lane that already contains a unit spawned on the previous timestep. \\
        \bottomrule
    \end{tabularx}
\end{table}

\subsection{Visualizations of Emergent Strategies}
\begin{figure}[h!]
    \centering
    \begin{subfigure}[b]{0.49\textwidth}
        \centering
        \includegraphics[width=0.8\linewidth]{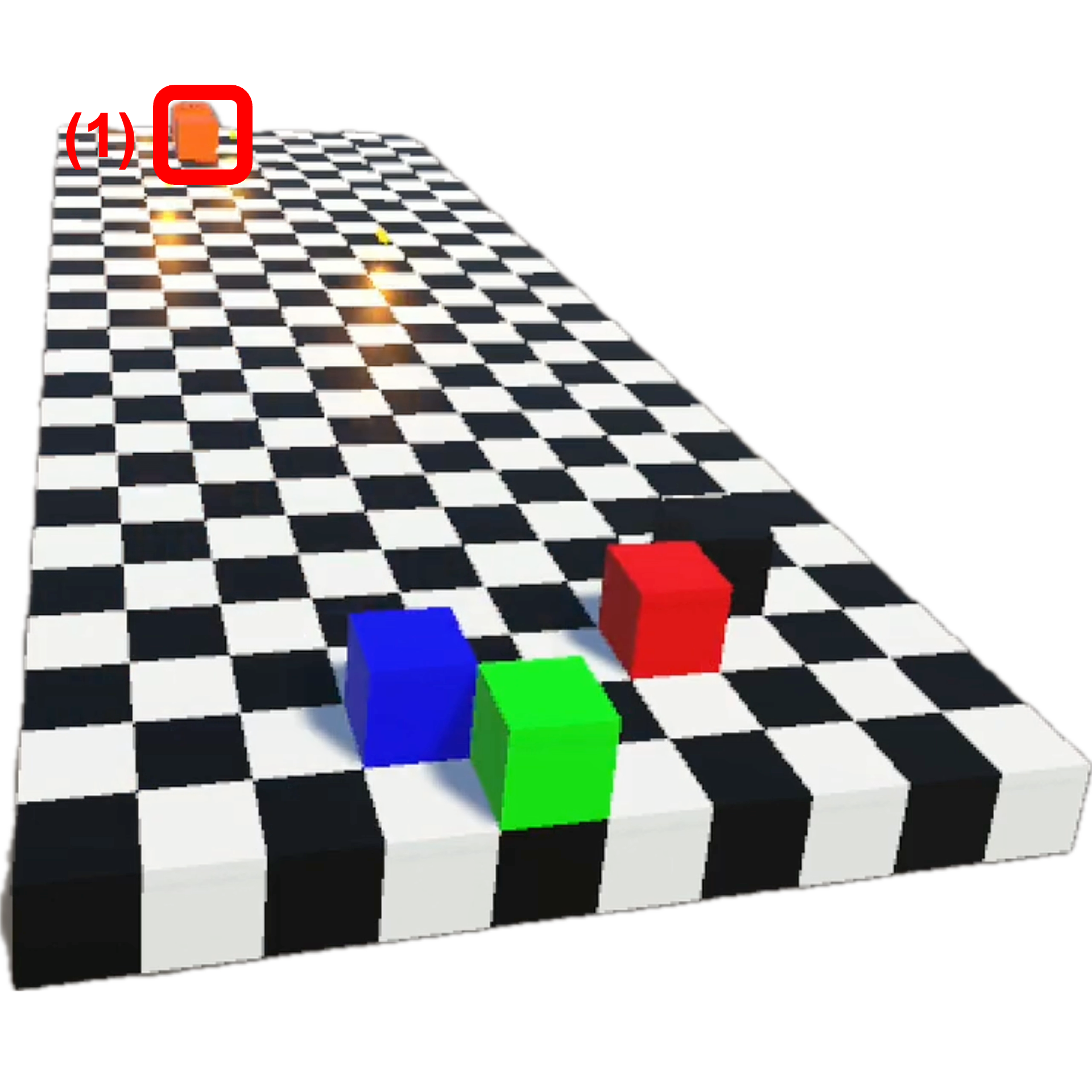}
        \caption{The Tandem Strategy. The attacker generates a durable unit followed immediately by a high-damage unit (1) in the same lane.}
        \label{fig:tandem}
    \end{subfigure}
    \hfill
    \begin{subfigure}[b]{0.49\textwidth}
        \centering
        \includegraphics[width=0.8\linewidth]{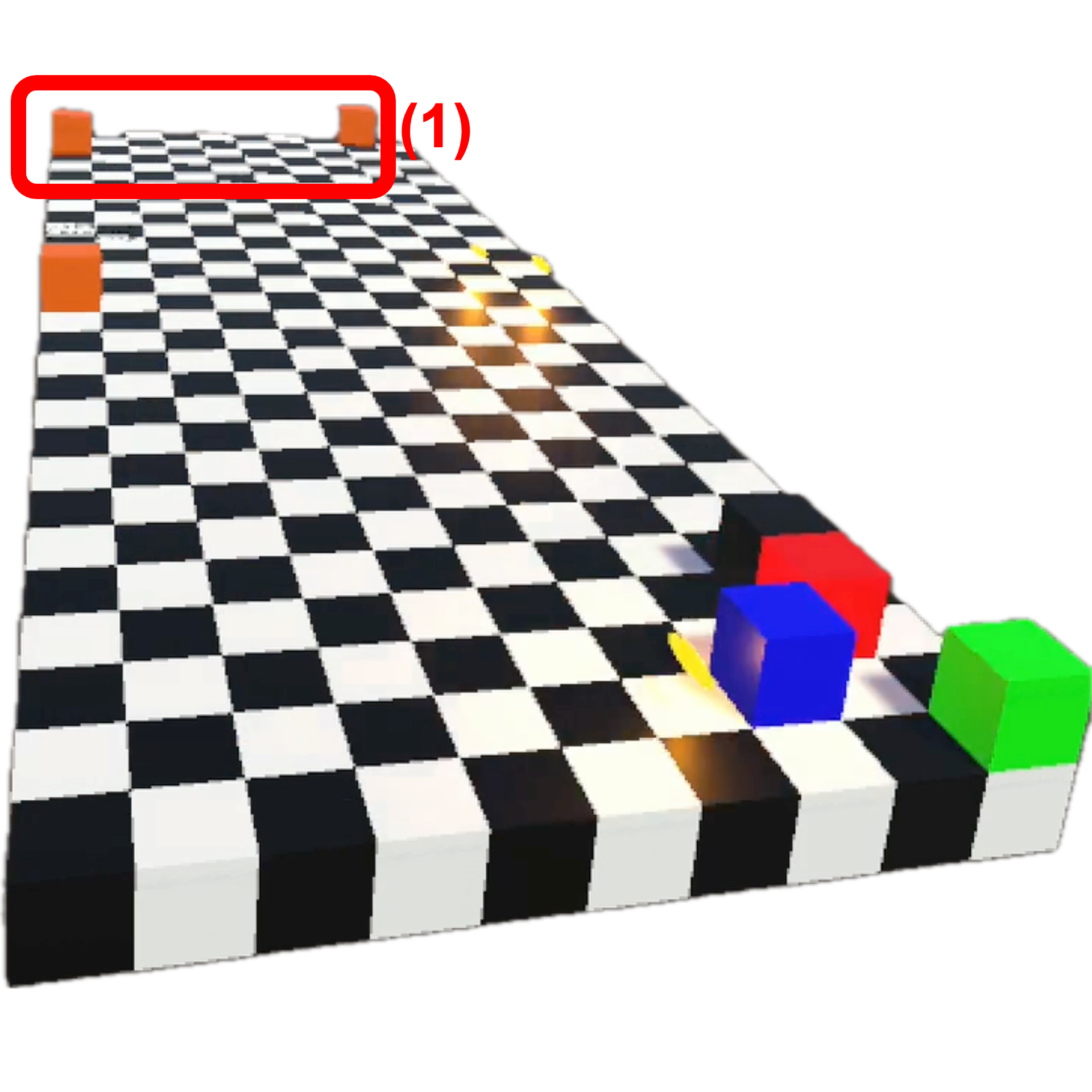}
        \caption{The Flanking Strategy. The attacker generates two threats (1) on opposite sides of the board simultaneously.}
        \label{fig:flanking}
    \end{subfigure}
    \caption{Examples of emergent adversarial strategies from the generative Attacker agent.}
    \label{fig:attacker_strategies}
\end{figure}

\begin{figure}[h!]
    \centering
    \begin{subfigure}[b]{0.49\textwidth}
        \centering
        \includegraphics[width=0.8\linewidth]{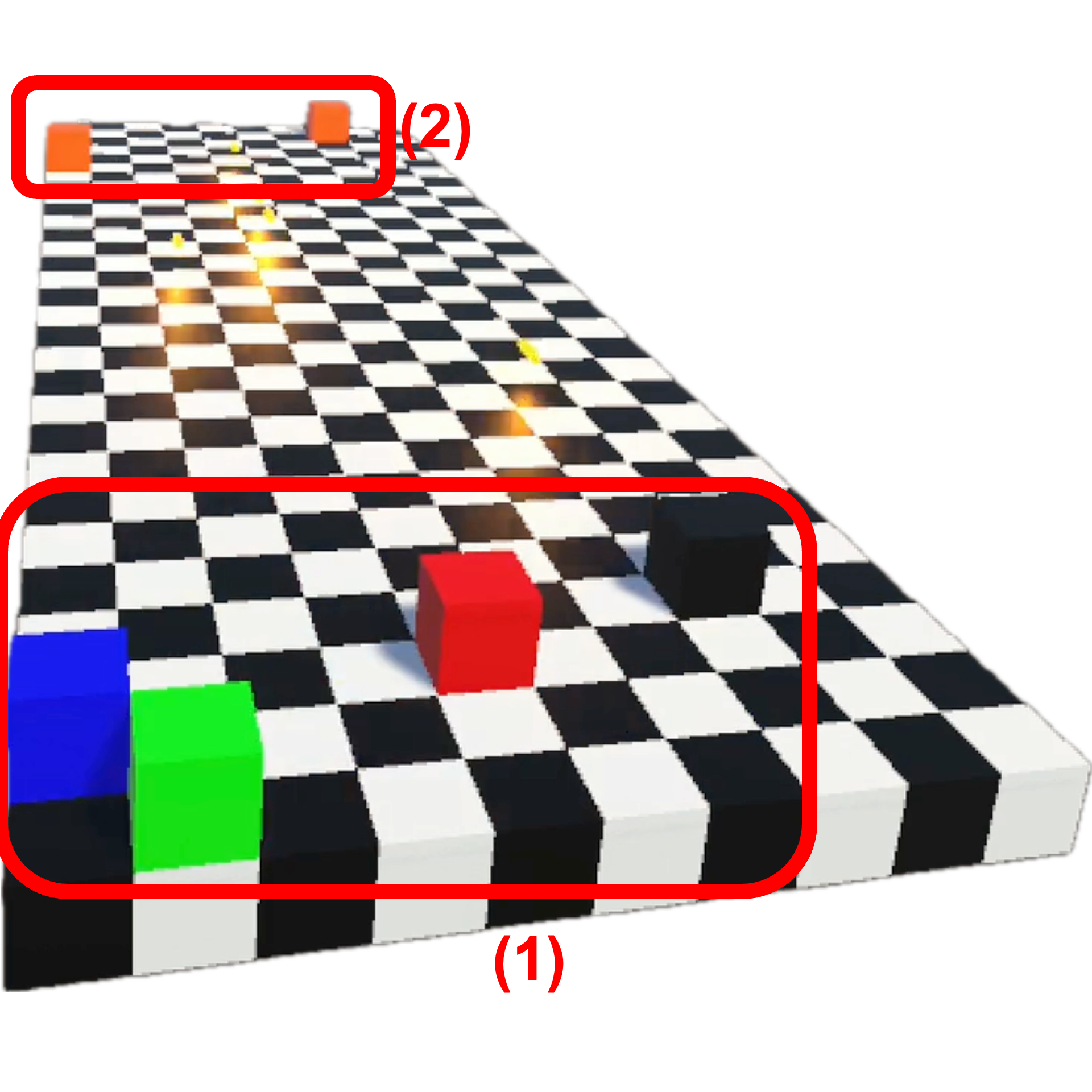}
        \caption{Cooperative Spreading. Defenders (1) position themselves in different lanes to counter multiple, spread-out units (2).}
        \label{fig:spread}
    \end{subfigure}
    \hfill
    \begin{subfigure}[b]{0.49\textwidth}
        \centering
        \includegraphics[width=0.8\linewidth]{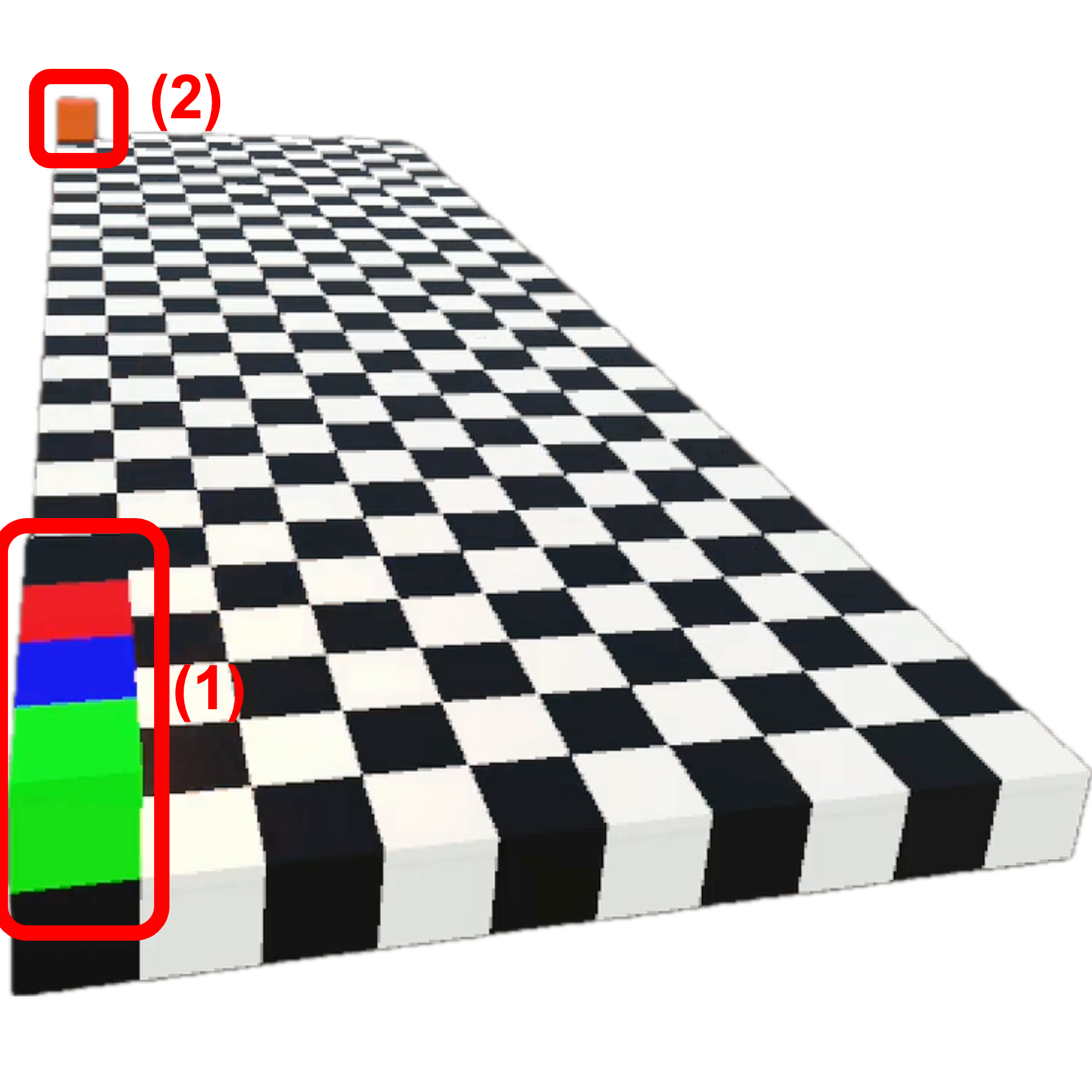}
        \caption{Cooperative Focusing. All defenders (1) converge on a single lane to focus fire on a high-priority target (2).}
        \label{fig:focus}
    \end{subfigure}
    \caption{Examples of emergent cooperative strategies from the Defender team.}
    \label{fig:defender_strategies}
\end{figure}

Figures \ref{fig:attacker_strategies} and \ref{fig:defender_strategies} provide visual examples of the key strategies that emerged during co-evolutionary training.

\end{document}